\documentclass[letterpaper]{article}
\usepackage{aaai24}  
\usepackage{times}
\usepackage{helvet}
\usepackage{courier}
\usepackage{xcolor}
\usepackage[hyphens]{url}
\usepackage{graphicx}
\usepackage{natbib}
\usepackage{caption}
\usepackage{multirow,multicol,siunitx}
\definecolor{cobalt}{rgb}{0.0, 0.28, 0.67}
\definecolor{darkorange}{rgb}{1.0, 0.55, 0.0}
\input{config.sty}
\title{Zero-shot Causal Graph Extrapolation from Text via LLMs}
\author{Alessandro Antonucci\thanks{Corresponding author: alessandro@idsia.ch.}, Gregorio Piqu\'e, and Marco Zaffalon}
\affiliations{IDSIA\\Lugano (Switzerland)}
\begin{document}
\maketitle
\begin{abstract}
We evaluate the ability of \emph{large language models} (LLMs) to infer causal relations from natural language. Compared to traditional natural language processing and deep learning techniques, LLMs show competitive performance in a benchmark of pairwise relations without needing (explicit) training samples. This motivates us to extend our approach to extrapolating causal graphs through iterated pairwise queries. We perform a preliminary analysis on a benchmark of biomedical abstracts with ground-truth causal graphs validated by experts. The results are promising and support the adoption of LLMs for such a crucial step in causal inference, especially in medical domains, where the amount of scientific text to analyse might be huge, and the causal statements are often implicit.
\end{abstract}

\section{Introduction}
In recent years, machine learning algorithms based on deep neural architectures have achieved astonishing results in many applied domains (see, e.g., \citeauthor{shen2017deep} \citeyear{shen2017deep} for an application in medical image analysis). These results are obtained by processing vast amounts of training data, allowing us to learn correlations and accurately solve predictive tasks. On the other side, scientific investigations need more than accurate predictions based on correlations, the focus being identifying \emph{causal} relations between the entities in the system under consideration. This requires dedicated formalisms and tools that cannot be provided by pure data-driven approaches, such as those considered by deep learning techniques \citep{bareinboim2022pearl}.

\begin{figure}[ht]
    \centering
    \includegraphics[width=8cm,height=4.6cm]{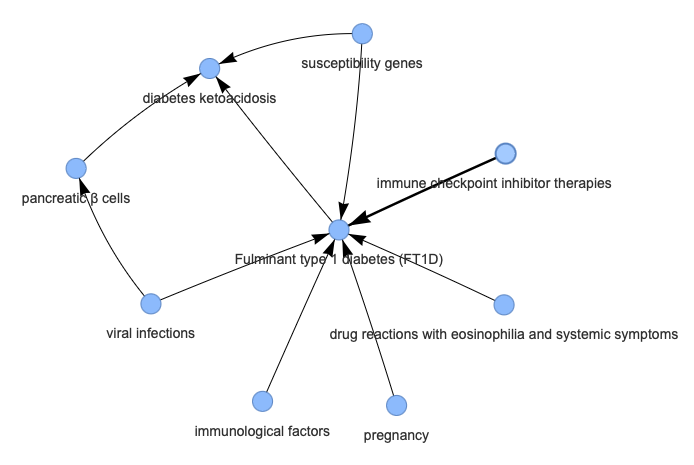}
    \caption{A (acyclic) CG modelling the causal relations included in the medical abstract in Figure~\ref{fig:text}.}
    \label{fig:cg}
\end{figure}

Following the Pearlian approach to causality \cite{pearl2009causality}, the first tool to go beyond a pure correlational analysis is to pair the observational data with a \emph{causal graph} (CG) modelling cause-effect relations as directed arcs connecting nodes associated with a system's entities. For example, Figure~\ref{fig:cg} depicts a CG for a medical domain. The classical \emph{do calculus} \cite{pearl2009causality} allows to compute (some) interventional queries from the CG and a set of observations. If knowledge about the structural models underlying the CG is also available, even the more challenging counterfactual queries can be addressed (e.g., \citeauthor{zaffalon2023approximating} \citeyear{zaffalon2023approximating}).

Randomised experiments are the traditional way to discover causal relations, hence CGs. Yet, such studies might be too expensive, time-consuming, or even impossible in many cases. This motivates \emph{causal discovery} intended as revealing causal information from purely observational data. Popular approaches are the \emph{Fast Causal Inference} by \citet{spirtes2000causation} based on a constraint-based procedure checking independence and the score-based \emph{Greedy Equivalence Search} by \citet{chickering2002optimal}. Yet, the output of these algorithms is generally a class of CGs, equivalent under the Markov condition, thus only partially solving the task. To consistently obtain a single CG, the only alternative not involving dedicated randomised experiments is to elicit the direction of the causal relations from a domain expert.

Especially for scientific and medical domains, the interaction with experts might also be indirect and consist of analysing the scientific literature on the topic and trying to grasp the causal relations stated in the text. Consider, for instance, the medical abstract in Figure~\ref{fig:text}. The text contains several causal statements summarised by the CG in Figure~\ref{fig:cg}. This paper focuses on automating such a \emph{natural language understanding} (NLU) task. 

\begin{figure}[ht]
\centering
\begin{minipage}{7cm}\scriptsize
Fulminant type 1 diabetes (FT1D) is a novel type of type 1 diabetes that is caused by extremely rapid destruction of the pancreatic $\beta$ cells. Early diagnosis or prediction of FT1D is critical for the prevention or timely treatment of diabetes ketoacidosis, which can be life-threatening. Understanding its triggers or promoting factors plays an important role in the prevention and treatment of FT1D. In this review, we summarised the various triggering factors of FT1D, including susceptibility genes, immunological factors (cellular and humoural immunity), immune checkpoint inhibitor therapies, drug reactions with eosinophilia and systemic symptoms or drug-induced hypersensitivity syndrome, pregnancy, viral infections, and vaccine inoculation. This review provides the basis for future research into the pathogenetic mechanisms that regulate FT1D development and progression to further improve the prognosis and clinical management of patients with FT1D.
\end{minipage}\caption{The medical abstract of \citet{luo2020fulminant}. Some cause-effect relations are highlighted (cause in blue, effect in orange).}
\label{fig:text}
\end{figure}

Even if no specific studies for the causal case are available (see the next section about the related work), classical and deep learning techniques for NLU have already been proven to classify relations among entities accurately \cite {li2022survey}. Those are typically supervised approaches requiring annotated data for the learning phase. This point might be critical because of the potential costs of the annotation process and the possible lack of diversity in the data.

Such a difficulty can be partially solved by using \emph{large language models} (LLMs), trained in an unsupervised manner from vast volumes of textual data. Here we consider the \emph{Generative Pre-trained Transformer} (GPT) language models designed to understand and generate human-like text based on the prompts. As the focus is on scientific domains, particularly medicine, we are not interested in the background, ``commonsense'', knowledge embedded in the LLM. By standard prompting techniques, we force the answer to the query to be only based on the source document. We do not expect to discover new knowledge this way but rather combine and standardise already existing information.

The paper is organised as follows. First we discuss the related work and notice how the direction we are considering, i.e., the unsupervised, zero-shot, extrapolation CGs from natural text through LLMs is relatively unexplored. Thus, to advocate our point, after a discussion on the necessary prompting techniques, we first consider a benchmark of pairwise relations and we test the accuracy of GPT in the causal recognition task against the existing results in the literature. The good performances we observe motivate us to extend the same technique to GCs. Finally we discuss the necessary future work required by this preliminary-but-promising analysis.

\section{Related Work}
The problem we consider can be intended as a (causal) specialisation of learning a \emph{knowledge graph} from natural language in a purely unsupervised manner. The literature for such a general task is vast (see \citeauthor{ji2021survey} \citeyear{ji2021survey} for a recent survey) and also includes recent attempts based on LLMs \cite{pan2023large}. Yet, the particular case of CGs received relatively little attention and, to the best of our knowledge, the only work in the same direction we consider is by \citet{arsenyan2023large}. This is an interesting preliminary study adopting the BERT-based models. Yet, it misses ground-truth data, thus not offering any quantitative baseline for our study.

From a more general perspective, the empirical analysis of the potential of LLMs in causal inference is attracting growing interest in the recent literature. Causal inference datasets in NLP primarily rely on discovering causality from empirical knowledge (e.g. commonsense knowledge), while we are interested in specific, scientific, knowledge from selected sources. 
Moreover, the work by \citet{jin2023can} outlines a key shortcoming of LLMs in terms of their causal inference skills, and show that these models achieve almost close to random performance on the task. Following \citet{zhang2023understanding} LLMs are not yet able to provide satisfactory answers for discovering new causal knowledge or for high-stakes decision-making tasks with high precision. 

\section{Prompt Engineering}
LLMs are known to often achieve satisfactory results in question answering (see, e.g., \citeauthor{singhal2023towards} \citeyear{singhal2023towards}). However, some expedients have been shown to increase the accuracy of queries. These techniques, known as \emph{prompt engineering} \cite{reynolds2021prompt}, can be regarded as rules and instructions to enhance the LLM capabilities on various tasks. Prompt engineering involves crafting systems and user messages that guide the LLM responses and shape its output to meet specific requirements. 

Among the many prompt engineering techniques, improving the clarity and precision of the prompt text by providing precise and specific instructions represents the most obvious strategy. Delimiters like brackets, tags or quotes can separate sections within the prompt, aiding in a more organised interpretation of the input. Furthermore, prompting for structured output by specifying the desired response format guides the model in generating well-organised results. Checking task conditions also ensures that the necessary assumptions are met. For instance, the prompt can verify whether essential information is available to complete the task and provide alternative instructions if this information is missing.

\floatstyle{boxed}
\restylefloat{figure}
\begin{figure}[ht]
\centering
\tiny
\begin{verbatim}
You will be provided with a text delimited by the <Text></Text> 
xml tags, and a pair of entities delimited by the <Entity></Entity>
xml tags representing entities extracted from the given text.

  Text:
   <Text>Cobalt metal fume and dust cause upper respiratory
   tract irritation, chronic interstitial pneumonitis,
   and skin sensitization.</Text>

  Entities:
   <Entity>fume</Entity>
   <Entity>sensitization</Entity>

Read the provided text carefully to comprehend the context 
and content. Examine the roles, interactions, and details 
surrounding the entities within the text.

Based only on the information in the text, determine the
most likely cause-and-effect relationship between the entities 
from the following listed options (A, B, C):
                        
  Options:
   A: "fume" causes "sensitization";
   B: "sensitization" causes "fume";
   C: "fume" and "sensitization" are not directly causally related;

Your response should analyze the situation in a step-by-step manner, 
ensuring the correctness of the ultimate conclusion, 
which should accurately reflect the likely causal connection 
between the two entities, based on the information
presented in the text. If no clear causal relationship is apparent, 
select the appropriate option accordingly.

Then provide your final answer within the tags
<Answer>[answer]</Answer>, (e.g. <Answer>C</Answer>).
\end{verbatim}
\caption{GPT prompt for causal relation discovery.}\label{fig:gpt-prompt}
\end{figure}

The technique called \emph{few-shot prompting} instead involves showing successful task examples to the model before requesting similar ones. This helps the model understand the context better, preparing it to deliver pertinent and accurate responses. Another principle of prompt engineering is giving the model enough time to ``think''. This involves requesting the model to answer with a step-by-step explanation of its thought process before providing the final answer. Doing so allows the model to work out its solution rather than rushing to conclusions. This principle applies, for example, when the model needs to verify the accuracy of a given solution. In this case, it is prompted to formulate its solution and then compare it to the one provided.

In the experiments discussed in the next two sections, we used GPT-4 Turbo as the LLM of choice for our study. The interaction with the LLM was achieved through a Python API\footnote{\url{https://platform.openai.com/docs/api-reference/chat}.} in a AMD EPYC 7742 server (2.25-3.4GHz, 128GB). Our code is freely available in a repository, where all the details about our prompt engineering choices can be retrieved.\footnote{\url{https://github.com/IDSIA-papers/causal-llms}.} Some of our design strategies can be seen in Figure~\ref{fig:gpt-prompt}, where we show the prompt used to query the LLM for discovering the causal relationship between a pair of entities extracted from a natural language text. Additional prompts used in our experiments are available in the repository mentioned above. These include prompts designed to explain the reasons behind the model's chosen answers or correct contradictions the LLM made with its answers.

\section{Learning Pairwise Relations}
We first test the ability of GPT to recognise the right ``orientation'' in a causal relation. In practice, given a sentence where the two relevant entities are identified, assuming we know that a direct cause-effect relation links these, we want to decide whether the first entity causes the second or vice versa.

For an empirical evaluation, we consider the dataset from \citet{hendrickx2019semeval} used for a competition of the popular \emph{SemEval} workshops.\footnote{\url{https://semeval.github.io}.} The dataset includes \num{10217} sentences. Each sentence has tags to identify two relevant entities and an annotation about the kind of relation and its orientation. The sentences with cause-effect relations are \num{1003}, while the others refer to other categories (e.g., member-collection, product-producer or content-container). Table~\ref{tab:pairs} reports a few examples of causal sentences with entity tags and orientation. 

\begin{table}[ht]
\centering
\begin{tabular}{p{6cm}c}
\hline
Sentence&Orientation\\
\hline
\hline
{\it \footnotesize \phantom{.}{\color{cobalt}{Zinc}} is essential for {\color{darkorange}{growth}} and cell division.}&{\footnotesize {\color{cobalt}{A}} $\to$ {\color{darkorange}{B}}}\\ 
{\it \footnotesize The {\color{cobalt}{infection}} came from a {\color{darkorange}{wound}}.}&{\footnotesize {\color{cobalt}{A}} $\leftarrow$ {\color{darkorange}{B}}}\\
{\it \footnotesize As we saw earlier, {\color{cobalt}{helicobacter}} is responsible for causing {\color{darkorange}{stomach ulcer}}.}&{\footnotesize {\color{cobalt}{A}} $\to$ {\color{darkorange}{B}}}\\ 
{\it \footnotesize The {\color{cobalt}{pseudolesion}} was caused by {\color{darkorange}{drainage}} of the paraumbilical vein.}&{\footnotesize {\color{cobalt}{A}} $\leftarrow$ {\color{darkorange}{B}}}\\
\hline
\end{tabular}
\caption{Four examples of relations in the benchmark dataset. The cause-effect orientation between the first (blue) and the second (orange) entity is also reported.}
\label{tab:pairs}
\end{table}

We ask GPT to detect the right orientation among the two entities by a prompt analogous to that in Figure~\ref{fig:gpt-prompt}. Our prompt needs an average time of $11.5 \pm 3.8$s to obtain an anwser about the orientation. The LLM might also deny the presence of a causal relation. Yet, this is the case only for five sentences ($\simeq 0.4\%$ of the cases). Table~\ref{tab:confusion} depicts the confusion matrix for the orientation predictions in the remaining \num{998} cases. The very accurate performance (F1-score $\simeq 99\%$) advocates the ability of GPT in properly recognising the right orientation in a causal sentence. 

\begin{table}[ht]
\centering
\begin{tabular}{l|l|c|c|c}
\multicolumn{2}{c}{}&\multicolumn{2}{c}{Ground Truth}&\\
\cline{3-4}
\multicolumn{2}{c|}{}&{\color{cobalt}{A}} $\to$ {\color{darkorange}{B}}& {\color{cobalt}{A}} $\leftarrow$ {\color{darkorange}{B}}\\
\cline{2-4}
\multirow{2}{*}{GPT}& {\color{cobalt}{A}} $\to$ {\color{darkorange}{B}} & $335$ & $7$\\
\cline{2-4}
& {\color{cobalt}{A}} $\leftarrow$ {\color{darkorange}{B}} & $6$ & $650$\\
\cline{2-4}
\end{tabular}
\caption{Confusion matrix for cause-effect orientation.}\label{tab:confusion}
\end{table}

Notably, a direct inspection of the thirteen misoriented sentences highlights possible errors in the annotation of the SemEval data set. Some examples are reported in Table~\ref{tab:wrong_pairs}.

\begin{table}[ht]
\centering
\begin{tabular}{p{5cm}c}
\hline
Sentence&(Mis)Orientation\\
\hline
\hline
{\it \footnotesize \phantom{.}{\color{cobalt}{Alternators}} generate {\color{darkorange}{electricity}} by the same principle as DC generators.}&{\footnotesize {\color{cobalt}{A}} $\leftarrow$ {\color{darkorange}{B}}}\\ 
{\it \footnotesize The {\color{cobalt}{movement}} developed from the {\color{darkorange}{rediscovery}} by European scholars of many Greek and Roman texts.}&{\footnotesize {\color{cobalt}{A}} $\to$ {\color{darkorange}{B}}}\\ 
{\it \footnotesize The {\color{cobalt}{cow}} makes a {\color{darkorange}{sound}} called lowing, also known as mooing.}&{\footnotesize {\color{cobalt}{A}} $\leftarrow$ {\color{darkorange}{B}}}\\ 
{\it \footnotesize Defra identified the different {\color{cobalt}{noises}} made by {\color{darkorange}{dogs}} and the meanings behind them.}&{\footnotesize {\color{cobalt}{A}} $\to$ {\color{darkorange}{B}}}\\ 
{\it \footnotesize The relative {\color{cobalt}{calm}} produced by the Shia {\color{darkorange}{ceasefire}} has coincided with what the CIA is now calling the "near strategic defeat" of al-Qaeda in Iraq.}&{\footnotesize {\color{cobalt}{A}} $\to$ {\color{darkorange}{B}}}\\ 
{\it \footnotesize The backup {\color{cobalt}{vocals}} are from a rather talented {\color{darkorange}{female}}, Stephanie Eitel.}&{\footnotesize {\color{cobalt}{A}} $\to$ {\color{darkorange}{B}}}\\ 
\hline
\end{tabular}
\caption{Wrong orientations detected by GPT.}
\label{tab:wrong_pairs}
\end{table}

The sentences in the dataset are already split in training and test set. For a comparison with other baselines, we consider the \num{2717} test sentences and extract the \num{325} causal relations. On those sentences only, our F1-score $\simeq 99\%$ is higher than the value obtained by the winner of the competition on the causal category (F1-score $\simeq 90\%$, as reported by \citet{hendrickx2019semeval}). Notably such a good performance has been achieved without the \num{8000} training samples used for the training by the other method. Similar considerations can be done for a comparison against the deep learning approaches considered by \citet{yin2017comparative}. 

The good performance achieved in the orientation task even suggest using LLMs as a post-processing for the output of the sets of Markov equivalent CGs returned by causal learning algorithms when both data and texts are available.

\section{Causal Graph Extrapolation}
The pairwise procedure discussed in the previous section can be naturally extended to CG extrapolation by iterated applications on all the possible pairs of entities. We consider a benchmark of \num{20} medical papers to be processed by the repeated pairwise procedure for preliminary validation. We restrict our attention to the abstracts, regarded here as a reliable summary of the causal relations discussed in the paper. 

The LLM is also used for \emph{entity recognition}. For this task, the medical text is complemented with additional information about the types of entities to be extracted. Since we focus on medical literature, the model was explicitly instructed to identify entities with a particular emphasis on diseases, medications, treatments, and symptoms. Additional operations are performed to recognise synonyms, redundant entities, or entities and names that can be used interchangeably. In the generated output, entities with synonymous or similar meanings are matched. The relatively small number of entities ($\leq$\num{20}) extracted from each abstract makes it possible to execute the, quadratic-time, iterated pairwise approach in a limited amount of time ($\leq$\num{30} minutes per abstract).

We denote the resulting CG as LLM-CG. CG$^*$ is instead the ground-truth CG obtained by asking domain experts to highlight the causal relations in the abstract.  Note that the entities the experts consider during their annotation are those returned by the LLM. In this setup, a false positive is an arc present in LLM-CG but not in CG$^*$, while a false negative is an arc missing in LLM-CG but not in CG$^*$.

\begin{figure}[ht]
    \centering
    \includegraphics[width=7cm]{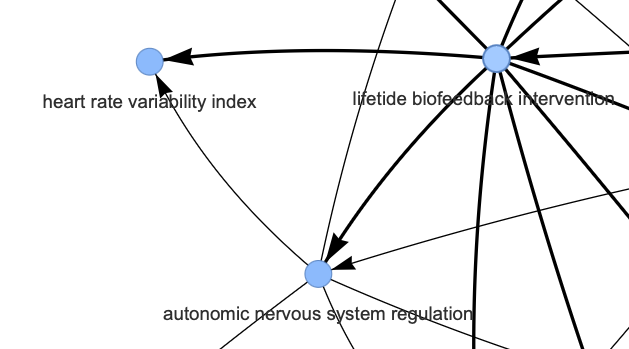}
    \caption{A multiply connected pattern in a CG.}
    \label{fig:multi}
\end{figure}

In the experiment we observe relatively few false negatives and hence a high recall $\simeq$ \num{97}$\%$, but, due to higher number of false positives a much lower precision $\simeq$ \num{74}$\%$. The result denote a good ability of LLMs in properly grasping the actual causal relations in the text. We conjecture that the high number of false positives might reflect the inability of the model LLM in distinguishing between direct cause-effect relations and indirect ones. Consider for instance the slice of CG in Figure~\ref{fig:multi}. Such a multiply connected topology might require an additional prompt engineering effort in order to decide whether the LLM properly grasps the direct influence of A toward C or should be instead regarded as a LLM reasoning based on the transitive property (A affect B and B affects C).

\begin{figure}[ht]
\centering
\begin{minipage}{7cm}\scriptsize
$[\ldots]$ Another patient had a significantly increased heart rate variability index without obvious changes in heart rate after the intervention. By reestablishing the balance in autonomic nervous system regulation and enhancing peripheral microcirculation, lifetide biofeedback intervention helps to maintain stable blood glucose levels, achieve disease remission $[\ldots]$
\end{minipage}\caption{An extract from the medical abstract inducing the multiply connected pattern in Figure~\ref{fig:multi}.}
\label{fig:text_multi}
\end{figure}

\begin{figure}[ht]
    \centering
    \includegraphics[width=6cm]{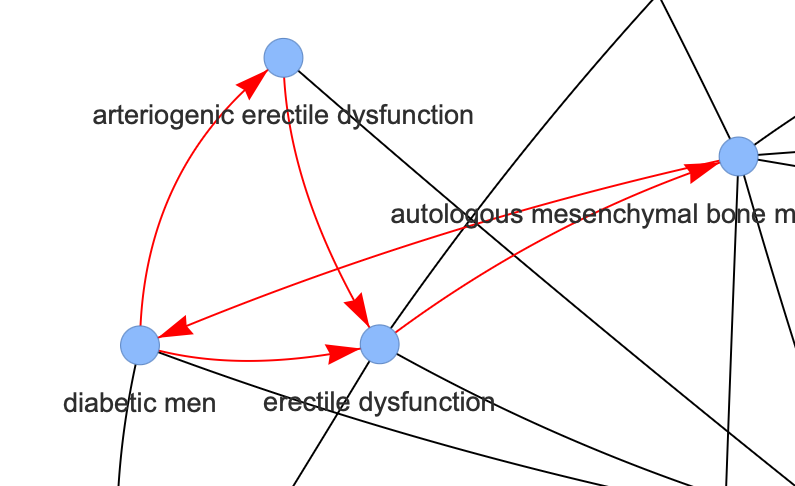}
    \caption{Two directed cycles in a CG.}
    \label{fig:multi2}
\end{figure}

In some cases, we also observed directed cycles in the resulting CG (Figure~\ref{fig:multi2}). Those cycles always have length bigger than two, i.e., we never observe two entities connected by arcs of opposite orientations. This was also the case for the pairwise experiments discussed in the previous section. Nevertheless, directed cycles might arise in causality to model special situations such as the presence of feedback loops \citep{rehder2017reasoning}. This is not the case of the medical domains under consideration for our benchmark texts, and the in fact the ground-truth CGs are all acyclic. If a cyclic CG arises in a domain that seems to contradict such a possibility, we might again design dedicated prompt engineering approaches to for the model acyclicity.

\section{Conclusions and Future Work}
We presented the results of some preliminary experiments performed with GPT for CG extrapolation from text. A number of specific prompt engineering strategies are also discussed. The results are promising, especially concerning the orientation of pairwise relations already classified as cause-effect. We consequently regard our approach as a natural post-processing tool to be used to refine the output of standard algorithms for causal discovery when training data are available together with text. If only text is available, the LLM might add to the CG arcs that are not reflecting a direct cause-effect relation.

As a future work we should intend to significantly expand the benchmark of ground-truth CGs for a deeper validation and extend the analysis to hybrid approaches mixing LLMs processing data and causal discovery algorithms processing data.
  opportunities.

\end{document}